# Towards a new system for drowsiness detection based on eye blinking and head posture estimation

Mejdi Ben Dkhil, Ali Wali, and Adel M.Alimi

*REsearch Group on Intelligent Machines*
*University of Sfax, National School of Engineers (ENIS)*
*BP 1173, 3038*
*Sfax, Tunisia*
*{mejdi.bendkhil, ali.wali, adel.alimi}@ieee.org*

*Abstract*—Driver drowsiness problem is considered as one of the most important reasons that increases road accidents number. We propose in this paper a new approach for real-time driver drowsiness in order to prevent road accidents. The system uses a smart video camera that takes drivers faces images and supervises the eye blink (open and close) state and head posture to detect the different drowsiness states. Face and eye detection are done by Viola and Jones technique.

*Keywords-component; Viola and Jones, drowsiness, eye blink, head posture.*

## I. INTRODUCTION

Driver drowsiness is a major factor for the increasing number of vehicles accidents. Klamer et al. [1] has shown that driving drowsiness increase from four to six times the risk of having an accident. Indeed, according to [2], driver drowsiness presents about 20% from all accident.

With the development of technology, various studies and techniques were proposed in order to acquire a real-time controlling drowsiness state and warming driver in order to prevent road accidents.
For this serious risk, we propose to develop a system which controls the driver drowsiness states in a real-time, and alerts driver in critical moment (feeling exhausted) in order to prevent accidents.

The first part of the paper is based on six major points as follows. The second section reviews some related works dealing with the face expressing recognition while the third deals with the formulation problems and challenges of learning the complexity of drowsiness detection. Besides, the fourth describes briefly our new approach and while in the fifth part, we will discover the experimental results in this work. Finally, the conclusion and some suggests for future works.

## II. RELATED WORKS

In this article, we are focusing on previous works to determinate the detection of drowsiness. To note, eye detection refers always to face detection.

Eye detection and tracking has been considered as one of the advanced techniques used for human computer interaction. This technique has developed many axes of researches, especially in our axis of work, all systems that have been developed to monitor driver drowsiness. It controls eye blink, eye movement, eye tracking. Various approaches are being used currently for face and eye detection. Those approaches can be divided into three main categories [3]:

- The first category includes methods based on biomedical signals [4].

- The second category includes methods based on driving behavior [5,6].

- The third category based on computer vision which has been the natural technique for controlling driver's state from face images.

According to Garcia et al. [7], the first category uses electrodes attached to the body which can be always bothering to driver. The second category requires many long training period while the third one is based on visual assessment.

This paper presents a new system for monitoring driver drowsiness cases based on computer vision techniques.

## III. PROBLEM FORMULATION

For eye detection, we always need face detection. In fact, it is worth noting that despite the important progress which has been made, detecting head pose and eye blinking with a high accuracy remains difficult due to the complexity of facial expressions.

The literature on facial expression recognition in static images is somewhat sparse in comparison with that of face recognition. Most of the existing references contain algorithms to extract features from an image and to reduce the dimensionality of the problem. Indeed the main problems are in particular the wide range of faces from one person to another and the vast range of possible facial expressions, variable timing and appearance both by age, gender, ethnicity, hair appearance, and the presence of various accessories (glasses, locks). Added to this, we cite the difficulties of shooting, such as changes in head position, partial lighting, and occlusions. Finally, the technical constraints imposed to limit the required memory space and computing time to allow the execution in real time.

## IV. PROPOSED APPROACH

In our system, a smart camera has been attached on the dashboard of car. It takes images of different states of driver's drowsiness detection.

The system consists of three phases: face detection, eye detection and head posture estimation.

### A. Face and Eye detection

Face and eyes are detected by the method of Viola-Jones. This method allows the detection of objects for which learning was performed. It was designed specifically for the purpose of face detection, but may also be used for other types of objects. As a supervised learning method, the method of Viola-Jones requires hundreds to thousands of examples of the detected object to train a classifier. The classifier is then used in an exhaustive search of the object for all possible positions and sizes the image to be processed.

This method has the advantage of being effective, and rapid. The method of Viola-Jones uses synthetic representations of pixel values: the pseudo-Haar features. These characteristics are determined by the difference of sums of pixels of two or more adjacent rectangular regions (Fig 1.) For all positions in all scales and in a detection window, the number of features may then be very high. .i,e the best features are then selected by a method of boosting, which provides a "strong" classifier more by weighting classifiers "weak". The Viola-Jones method used by the Adaboost algorithm.

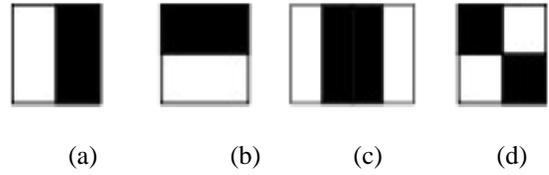

(a)    (b)    (c)    (d)

Fig1. Examples of neighborhoods used[8].

The exhaustive search for an object within an image that can be costly in computing time, the method detection organizes a cascade of classifiers, applied sequentially. Each classifier determines the presence or absence of the object in the image. The simplest and fastest classifiers are placed at the beginning that quickly eliminates many drawbacks.

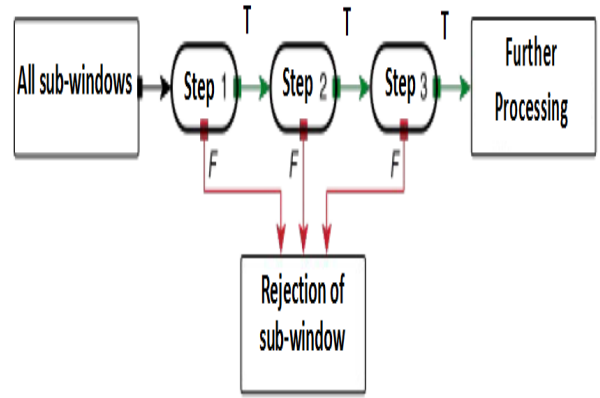

T: True
F: False

Fig2. Cascade of classifiers [8].

In general, the method of Viola-Jones provides good results in the
Face Detection or other objects, with few false positives for a calculation time quite low, allowing the operation here in real time.

### B. Eye Blinking

The detection of eye blinking in real time is very important to estimate driver drowsiness state.

In literature, the PERCLOS (Percentage of eye Closure) [10] value has been used as drowsiness metric which shows the percentage of closure in specific time (eg in a minute, eyes are 80% closed). Using these eyes closer and blinking ration, one can detect drowsiness of driver. Then, we move to the following frame until obtaining closed eyes.

We calculate the duration of eye closure; if it exceeds a predened Time T (for example 2 seconds), we can say that the driver enters in a drowsiness state, hence, our system launchs warning to wake the driver.

*C. Head posture estimation*

Estimating the pose of the head standardizes measures, also recalculate certain images by balancing rotations.

The proposed solution is to calculate the pitch which is defined by the angle of the eyes with the horizontal.

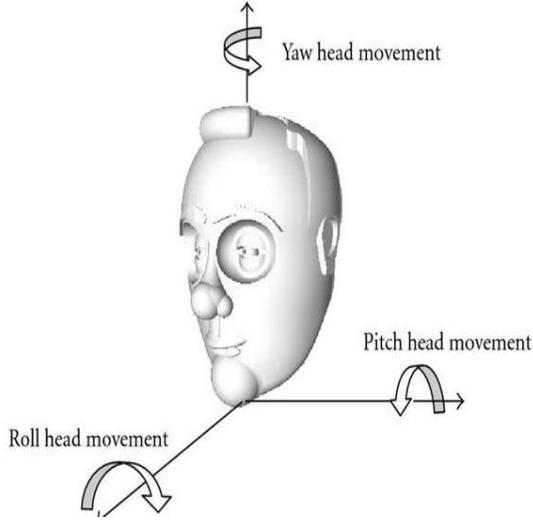

Fig3. Definition of head movements: pitch, roll, and yaw.

Three angles are considered to 0° when the head is opposite, no visible orientation relative to the camera. The relative position of the eyes, mouth, and temples allow finding the three angles of approximately. Yet, is it accurate enough to be useful?

These angle calculations are not affected by facial expressions, since the distances are constants used on the face and vary only with respect to the head position. In these calculations, the evaluation of the pitch only remains correct for large changes poses. The extracted sub-images can undergo against rotation on the pitch to allow greater precision in the measurements.

The evaluation of the pitch remains correct for large variations of poses.

$$\text{Pitch} = \arctan\left(\frac{easting\ Eyes}{Distance\ Eyes}\right) \quad (1)$$

In this work, we calculate the pitch head movement, if pitch=0°, the head is in his normal position, and if pitch ≠0°, we can know that the driver head's in movement.

*D. System overview*

The system architecture flowchart is shown in Fig4.

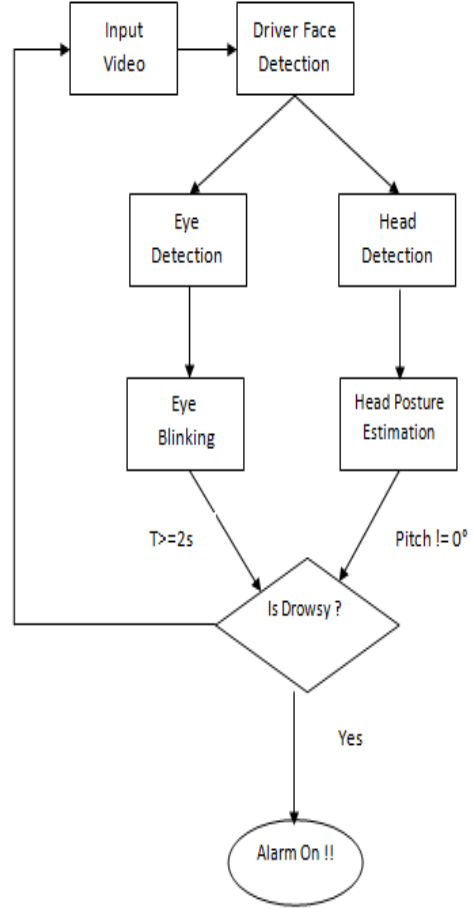

Fig4. Proposed approach for detecting driver drowsiness

We define three rules, to detect drowsiness state:

- **R1 :** *if* (T>2sec),

    Drowsiness=2.=> Red .

- **R2**: *if* ((T<2 sec )and (pitch = 1)),
    Drowsiness=1 => Yellow.

- **R3:** *if* ((T<3 sec) sec and (pitch=0)),

    Drowsiness =0 => Green.

If R1, the system alerts driver by a bip message "wake up", and a red LED lights up. If R2, a yellow LED lights up. And, if R3, we still in normal state, there is no risks.

## V. EXPERIMENTAL RESULTS

We are going to describe different experimental results as well as three approaches developed in this work.

### A. *Database*

The dataset consists of 1521 gray level images with a resolution of 384x286 pixels. Each one shows the frontal view of a face of one out of 23 different test persons. For comparison reasons, the set also contains manually eye positions. It also includes several appearances such as frontal face view, profile face view, faces with and without glasses.... This database is widely used for face and eyes detection. It has been recorded and published to give all researchers working in the area of face detection the possibility to compare the quality of their face detection algorithms with others [11].

### B. *Results*

- **Test1: Eye detection by haar classifiers**

We have implemented experiments on BioID database (Fig5.).The estimation of this algorithm is made by the calculation of the rate of good detections eye blink (GDR1) using the following formula.

$$GDR1 = \frac{\text{Number of detected eye}}{\text{Total eye number}} \quad (2)$$

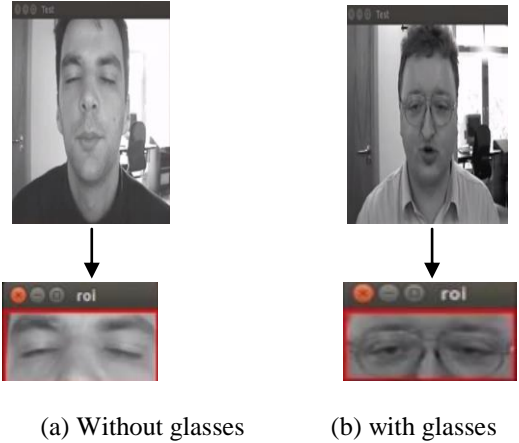

(a) Without glasses    (b) with glasses

Fig.5. Example of Eyes detection algorithm employing BioID database.

With this algorithm we achieved to detect 1442 eyes from 1521 with a 94.8% as GDR1.

**Performance of Haar classifiers in eyes detection (BioID Basis of faces)**

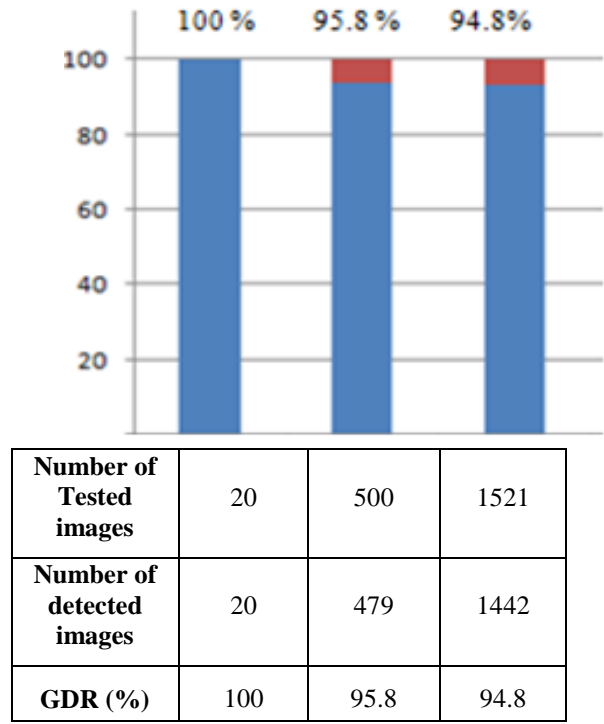

| Number of Tested images | 20 | 500 | 1521 |
|---|---|---|---|
| Number of detected images | 20 | 479 | 1442 |
| GDR (%) | 100 | 95.8 | 94.8 |

Fig.6. Eyes detection algorithm performances employing BioID basis.

- **Test2: head pose detection by haar classifiers**

We have implemented experiments on BioID database. The estimation of this algorithm is made by the calculation of the rate of good detections head posture (GDR2), using the following formula.

$$GDR2 = \frac{\text{Number of detected head pose}}{\text{Total images}} \quad (3)$$

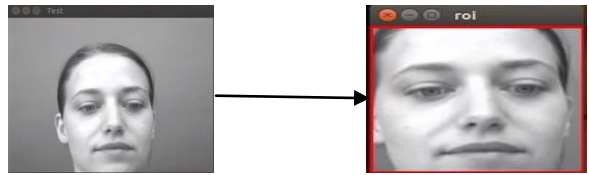

Fig.7. Example of heads poses detection algorithm employing BioID database.

With this algorithm we achieved to detect 1031 head poses from 1521 images with a 67.75% as GDR2 (Fig8).

- **Test3: Both Eye detection and Head pose by haar classifiers**

We have implemented experiments on BioID database. The estimation of this algorithm is made by the calculation of the rate of good detections head posture (GDR3), using the following formula.

$$GDR3 = \frac{\text{Number of detected eye and head pose}}{\text{Total images}} \quad (4)$$

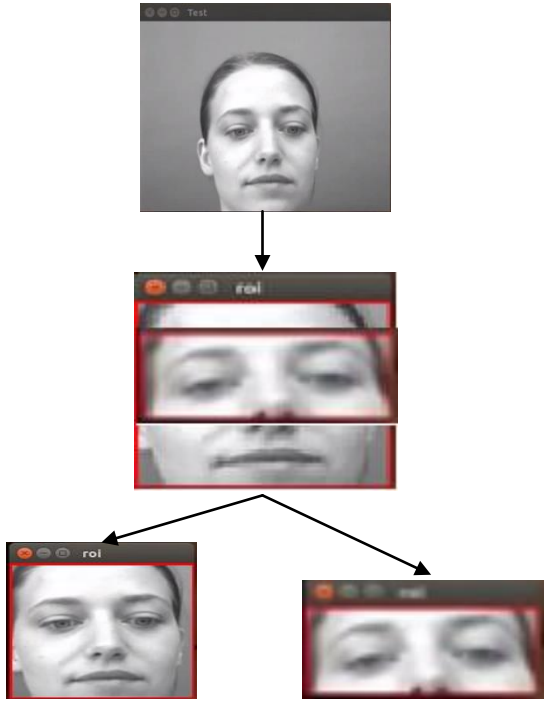

Fig.7. Example of Eyes and heads detection algorithm employing BioID database.

With this algorithm we achieved to detect 1263 head poses from 1521 images with a 83.03% as GDR3(Fig8).

Table1. Comparison of GDR given by different tests.

| Approach | Eye Blink | Head Posture Estimation | EYE BLINK and Head Pose |
|---|---|---|---|
| Number of detected images | 1442 | 1031 | 1263 |
| GDR (%) | 94.8 | 67.75 | 83.03 |

**Performance of Haar classifiers in different techniques detection (BioID Basis of faces)**

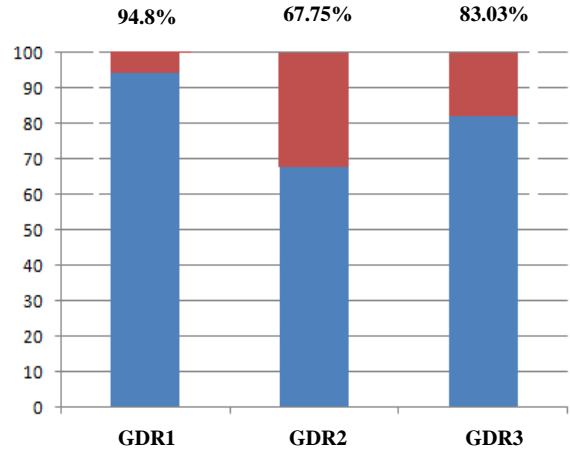

Fig8. Comparison of GDR given by different tests.

## VI. CONCLUSION AND FUTURE WORKS

We developed three approaches for driver monitoring drowsiness state in real time and we compared different results. As well, this paper presents a new approach for controlling driver drowsiness state in real time based on eye blinking and head posture. This developed system is based on computer vision techniques, and its robustagainst illumination variations, indeed it obeys real time computation needs.

Thus, the main goal of this project is to add brain and visual information, that both EEG and video methods are at last been combined to make the decisions in order to obtain a deeply reliable automatic drowsiness detector.

## VII. ACKNOWLEDGMENT

The authors acknowledge the financial support of this work by grants from General Direction of Scientific Research (DGRST), Tunisia, under the ARUB program.

## VIII. References

[1] S. G. Klauer, T. A. Dingus, V. L. Neale, , and J. D. Sudweeks, "The impact of driver inattention on near-crash/crash risk: An analysis using the 100-car naturalistic driving study data," *National Highway Traffic Safety Administration, DC, DOT HS*, vol. 810, 2006.

[2] J. Connor, R. Norton, S. Ameratunga, E. Robinson, I. Civil, R. Dunn, J. Bailey, and R. Jackson, "Driver sleepiness and risk of serious injury to car occupants: Population based control study." *British Medical Journal*, vol.

[3] L. M. Bergasa, J. Nuevo, M. A. Sotelo, R. Barea, and M. E. L. Guill´en,"Real-time system for monitoring driver vigilance," *IEEE Transactions on Intelligent Transportation Systems*, vol. 7, no. 1, pp. 63–77, 2006.

[4] C. Papadelis, Z. Chen, C. Kourtidou-Papadeli, P. Bamidis, I. Chouvarda, E. Bekiaris, and N. Maglaveras, "Monitoring sleepiness with on-board electrophysiological recordings for preventing sleep-deprived traffic accidents. "*Clinical Neurophysiology*, vol. 118, no. 9, pp. 1906–1922, September 2007.

.[5] T. Wakita, K. Ozawa, C. Miyajima, K. Igarashi, K. Itou, K. Takeda, and F. Itakura, "Driver identification using driving behavior signals," *IEICE - Trans. Inf. Syst.*, vol. E89-D, no. 3, pp. 1188–1194, 2006.

[6] Y. Takei and Y. Furukawa, "Estimate of driver's fatigue through steering motion." *IEEE International Conference on Systems, Man and Cybernetics.*, vol. 2, p. 1765–1770, 2005.

[7] I. Garcia,S. Bronte, L.M. Bergasa, J. Almazan and J. Yebes, "Vision-based drowsiness detector for Real Driving Conditions." Intelligent Vehicles Symposium, Spain, June 3-7,2012.

[8] P. Viola and M. J. Jones, "Robust real-time face detection," Int. J.Comput. Vision, vol. 57, no. 2, pp. 137–154, 2001.

[9] http://www.hindawi.com/journals/mse/2009/245606

[10] W.W. Wierwille, L.A. Ellsworth, S.S. Wreggit, R.J. Fairbanks and C.L.Kirn, "Research on vehicle based driver status/performance monitoring: development, validation and refinement of algorithms for detection of driver drowsiness," National Highway Traffic Safety Administration, Technical report, DOT HS 808 247, 1994.

[11] https://www.bioid.com/About/BioID-Face-Database.